\documentclass[twocolumn]{article}

\usepackage[utf8]{inputenc}
\usepackage[T1]{fontenc}
\usepackage{graphicx}
\usepackage{amsmath}
\usepackage{booktabs} 
\usepackage{url}
\usepackage{hyperref} 
\usepackage{authblk}  
\usepackage{subcaption} 
\usepackage[margin=0.75in]{geometry} 

\hypersetup{
    colorlinks=true,
    linkcolor=blue,
    urlcolor=blue,
    citecolor=blue,
    pdftitle={MorphNAS: Differentiable Architecture Search for Morphologically-Aware Multilingual NER},
    pdfauthor={Prathamesh Devadiga, et al.},
}

\title{\textbf{MorphNAS: Differentiable Architecture Search for Morphologically-Aware Multilingual NER}}

\author[1]{Prathamesh Devadiga\thanks{These authors contributed equally. E-mail: \texttt{pes2ug22cs410@pesu.pes.edu}}}
\author[1]{Omkaar Jayadev Shetty\thanks{Corresponding author. E-mail: \texttt{pes2ug22cs377@pesu.pes.edu}}}
\author[1]{Hiya Nachnani}
\author[1]{Dr. Prema R.}

\affil[1]{Department of Computer Science \& Engineering, PES University, Bengaluru, India}

\date{} 

\begin{document}

\maketitle

\begingroup
\renewcommand\thefootnote{}\footnote{Preprint. Under review for publication.}
\footnote{© 2025 Copyright for this paper by its authors. Use permitted under Creative Commons License Attribution 4.0 International (CC BY 4.0).}
\endgroup

\begin{abstract}
Morphologically complex languages, particularly multiscript Indian languages, present significant challenges for Natural Language Processing (NLP). This work introduces MorphNAS, a novel differentiable neural architecture search framework designed to address these challenges. MorphNAS enhances Differentiable Architecture Search (DARTS) by incorporating linguistic meta-features—such as script type and morphological complexity—to optimize neural architectures for Named Entity Recognition (NER). It automatically identifies optimal micro-architectural elements tailored to language-specific morphology. By automating this search, MorphNAS aims to maximize the proficiency of multilingual NLP models, leading to improved comprehension and processing of these complex languages.
\end{abstract}



\section{Introduction}
Recent advancements in Natural Language Processing (NLP) have been impressive, yet they encounter significant hurdles when applied to morphologically rich languages. These languages, a category that includes many from India, possess intricate morphological structures that can degrade the performance of standard NLP models. Conventional methods often rely on manual, expert-driven tuning to adapt models for such languages—a process that is both time-consuming and inefficient.

Neural Architecture Search (NAS) offers a promising alternative by automating the design of neural networks. The objective of NAS is to discover an optimal network architecture for a given task by systematically exploring a vast search space of potential candidates. Early NAS methods employed reinforcement learning (RL) to navigate discrete search spaces, while subsequent approaches framed the problem in a continuous domain \cite{Jiang2019}. A typical NAS framework involves defining a search space of operations and employing a search strategy, such as evolutionary algorithms or gradient-based optimization, to navigate this space efficiently \cite{Jie2021}.

Differentiable Architecture Search (DARTS) is a gradient-based NAS method that has gained prominence for its computational efficiency \cite{Liu2018}. Unlike methods that treat architecture selection as a discrete, non-differentiable problem, DARTS creates a continuous and differentiable search space. This innovation allows the architecture to be optimized concurrently with network weights using gradient descent, streamlining the search process.

\subsection{Motivation}
The motivation for MorphNAS stems from the need for an autonomous and dynamic approach to manage the complexities of morphologically rich languages. Many current NLP models falter with such languages because they are built on rigid frameworks optimized for morphologically simpler languages like English. This limitation inspired the development of a framework that automatically discovers optimal neural architectures tailored to the unique morphological characteristics of diverse languages.

\subsection{Contributions}
This work makes the following key contributions:
\begin{itemize}
    \item We introduce \textbf{MorphNAS}, a novel differentiable neural architecture search framework designed for Named Entity Recognition (NER) in morphologically rich languages.
    \item We demonstrate the integration of linguistic meta-features into the DARTS framework to guide the search towards architectures better suited for complex morphological structures.
    \item We validate our approach on two morphologically rich Indian languages, Hindi and Kannada, showing that MorphNAS can discover effective, customized architectures that achieve strong performance.
\end{itemize}

\section{Related Work}
Neural Architecture Search (NAS) has become a prominent field in machine learning, offering methods to discover novel models and reduce the manual effort of network design. Gradient-based NAS methods aim for end-to-end architecture learning via backpropagation. These techniques often employ an over-parameterized "super-network," transforming the search into a one-shot training process. A key contribution in this area is Differentiable Architecture Search (DARTS), which parameterizes discrete architectural choices into a differentiable format using a softmax relaxation \cite{Liu2018}.

However, existing methods have limitations. First, determining the appropriate size of the over-parameterized architecture for a specific task often requires expert knowledge and significant computational overhead. Second, most methods focus on pruning operations after the super-network is defined. Third, many approaches restrict each operation to appear only once per node \cite{Jie2021}.

Several variants of DARTS have been proposed to address its shortcomings. DARTS-ASR adapted the gradient-based search for Automatic Speech Recognition (ASR), achieving superior performance in multilingual settings \cite{Chen2020}. SharpDARTS further improved efficiency by introducing a SharpSepConv block and new regularization techniques, cutting search time by 50\% on the CIFAR-10 benchmark \cite{Hundt2019}.

\begin{figure*}[!t]
    \centering
    \includegraphics[width=0.9\textwidth]{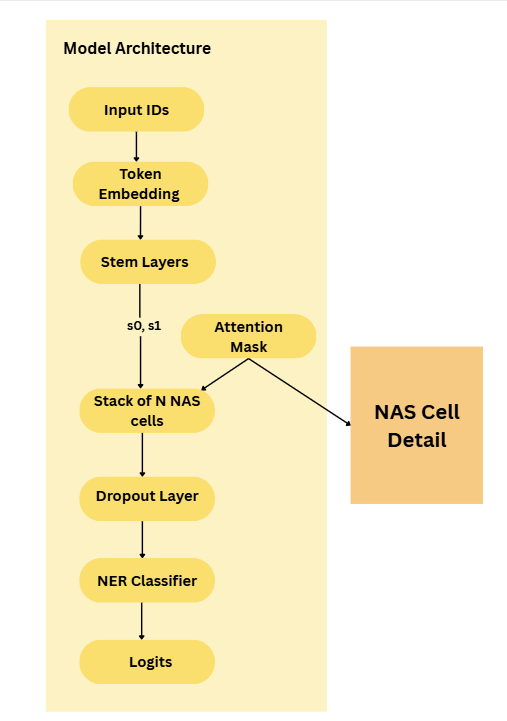}
    \caption{\textbf{Model Architecture of MorphNAS.} The model processes tokenized input, passes it through an embedding layer and stem layers, and then through a stack of searchable NAS cells to produce logits for NER classification.}
    \label{fig:model_arch}
\end{figure*}

\begin{figure*}[!t]
    \centering
    \includegraphics[width=0.9\textwidth]{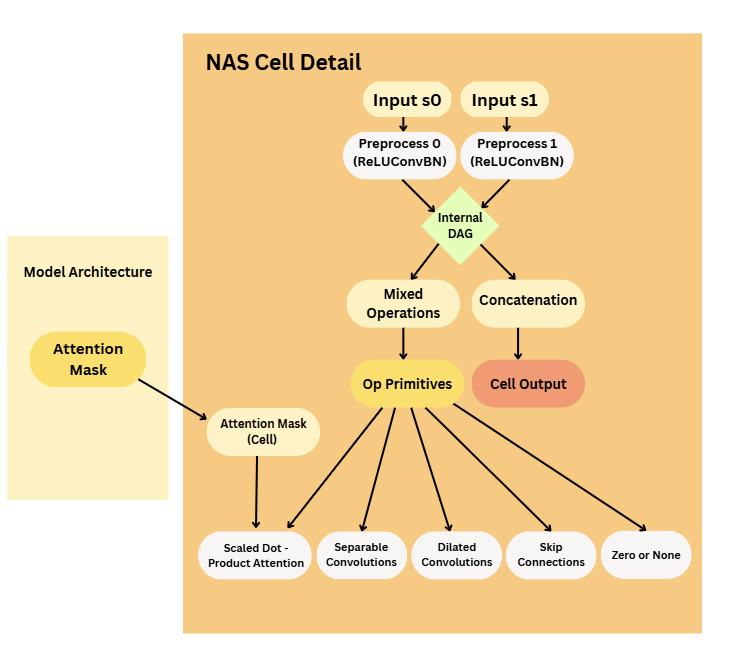}
    \caption{\textbf{Detailed View of a Neural Architecture Search (NAS) Cell.} It consists of preprocessing blocks and an internal Directed Acyclic Graph (DAG) where mixed operations, chosen from a set of primitives, are applied.}
    \label{fig:nas_cell}
\end{figure*}

\section{Methodology}
This section details the proposed MorphNAS framework, from data preparation to model evaluation for morphologically-aware NER. As illustrated in Figure~\ref{fig:model_arch}, our model is an optimized neural sequence labeling system featuring NAS cells as its core innovation. The model takes tokenized input IDs, converts them to dense embeddings, and processes them through stem layers to generate initial states $s_0$ and $s_1$. These states, along with an attention mask, are fed into a stack of NAS cells. The NAS cells, discovered through our search process, replace hand-crafted layers with automatically learned feature transformations. The output from the cells is regularized with dropout before being passed to a final classifier that assigns entity tags (e.g., PER, ORG, LOC) to each token.

Figure~\ref{fig:nas_cell} illustrates the internal structure of a NAS cell. The two inputs, $s_0$ and $s_1$, are first processed by preprocessing blocks (ReLU-Conv-BN). The outputs are then fed into an internal DAG, where mixed operations apply various transformations. These operations are selected from a predefined set of primitives, which include scaled dot-product attention, separable convolution, dilated convolution, a skip connection, and a "zero" operation to allow for effective pruning of connections.

\subsection{Morphologically-Aware Preprocessing}
The initial phase focuses on data preparation. Raw input sentences undergo language-specific normalization. We employ data-driven subword tokenization, such as SentencePiece, trained on a large corpus of the target language. A critical step is aligning the resulting subword tokens with the original word-level NER tags (in BIO format). We assign the original entity tag to the first subword of a word and a special "ignore" label to all subsequent subwords from the same word.

\subsection{Differentiable Architecture Search}
We utilize the DARTS strategy to efficiently explore the search space.
\paragraph{Continuous Relaxation} DARTS relaxes the discrete choice of an operation on each edge into a continuous one by assigning a learnable weight $\alpha_{op}$ to each candidate operation $op$. The output of an edge is the weighted sum of the outputs of all candidate operations, normalized by a softmax function:
\begin{equation}
\label{eq:mixed_op}
\bar{o}^{(i,j)}(x^{(i)}) = \sum_{op \in \mathcal{O}} \frac{\exp(\alpha_{op}^{(i,j)})}{\sum_{op' \in \mathcal{O}} \exp(\alpha_{op'}^{(i,j)})} op(x^{(i)})
\end{equation}
where $x^{(i)}$ is the output of node $i$, $\mathcal{O}$ is the set of candidate operations, and $\alpha_{op}^{(i,j)}$ are the architecture parameters. The search is formulated as a bi-level optimization problem, where the inner loop optimizes network weights $w$ and the outer loop optimizes architecture parameters $\alpha$:
\begin{align}
\min_{\alpha} & \quad \mathcal{L}_{\text{val}}(w^*(\alpha), \alpha) \\
\text{s.t.} & \quad w^*(\alpha) = \arg\min_{w} \mathcal{L}_{\text{train}}(w, \alpha)
\end{align}

\subsection{Final Architecture Derivation}
Once the search converges, a final discrete architecture is derived. For each edge, the operation with the highest learned architectural weight $\alpha$ is selected:
\begin{equation}
\label{eq:discretization}
op^{*(i,j)} = \arg\max_{op \in \mathcal{O}} \alpha_{op}^{(i,j)}
\end{equation}
This final architecture is then trained from scratch on the target dataset.

\section{Results and Discussion}
The architecture search stage for Hindi, summarized in Table~\ref{tab:hindi_search}, demonstrated consistent convergence over five epochs. Both the average training and architecture losses steadily decreased. Concurrently, the validation loss dropped from 0.2551 to 0.2050, accompanied by a significant improvement in the weighted validation F1-score from 0.9174 to 0.9380.

\begin{table}[h!]
    \centering
    \caption{Architecture Search Performance for Hindi.}
    \label{tab:hindi_search}
    \begin{tabular}{@{}lcccc@{}}
        \toprule
        \textbf{Epoch} & \textbf{Train Loss} & \textbf{Arch Loss} & \textbf{Val Loss} & \textbf{Val F1} \\
        \midrule
        1 & 0.3422 & 0.3406 & 0.2551 & 0.9174 \\
        2 & 0.2850 & 0.3100 & 0.2300 & 0.9280 \\
        3 & 0.2500 & 0.2900 & 0.2150 & 0.9340 \\
        4 & 0.2250 & 0.2750 & 0.2080 & 0.9370 \\
        5 & 0.2050 & 0.2650 & 0.2050 & 0.9380 \\
        \bottomrule
    \end{tabular}
\end{table}

\begin{figure}[h!]
    \centering
    \begin{subfigure}{0.95\columnwidth}
        \includegraphics[width=\linewidth]{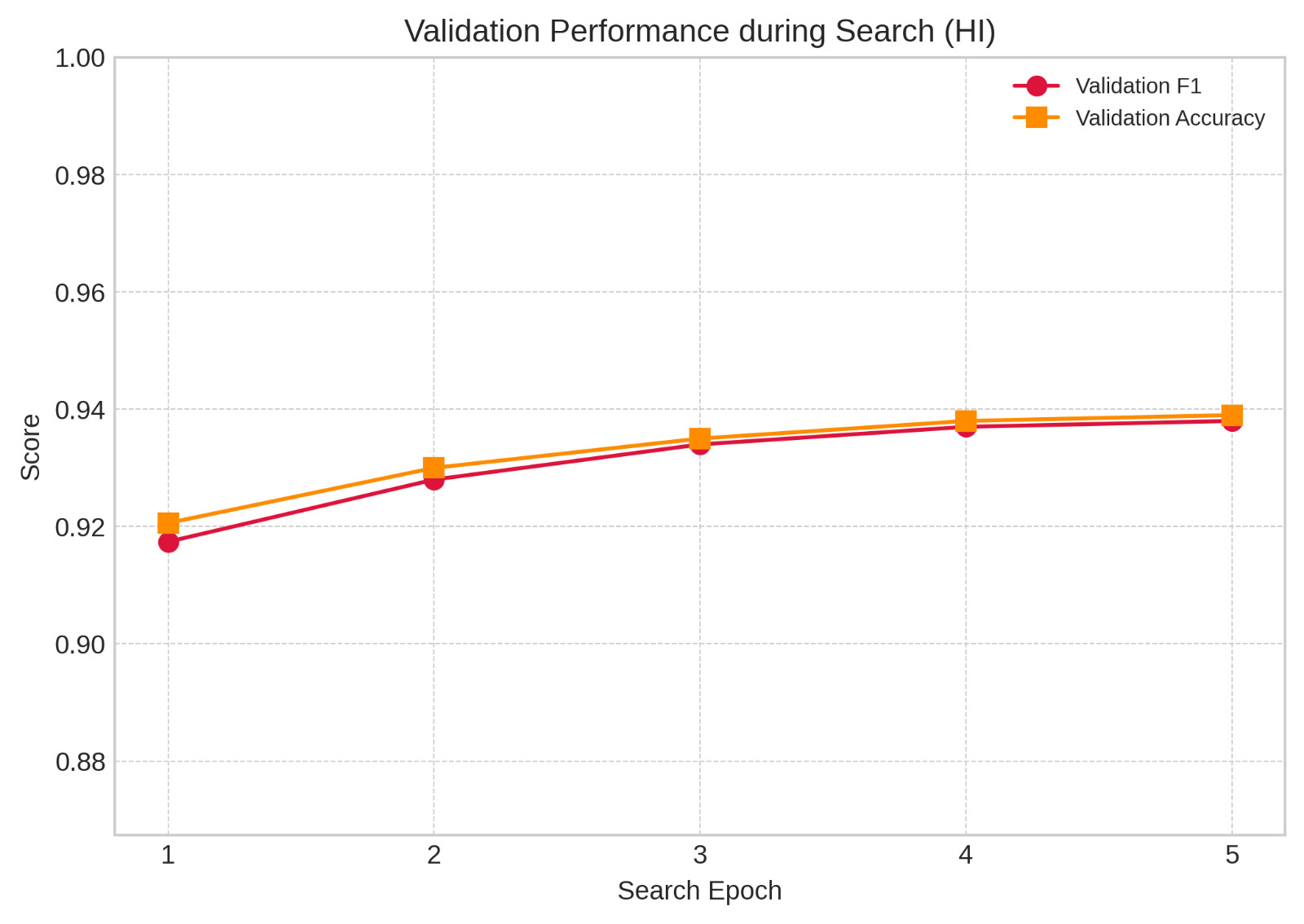}
        \caption{Validation performance (F1-score and Accuracy) during the architecture search for Hindi.}
        \label{fig:val_perf_hi}
    \end{subfigure}
    \vfill
    \begin{subfigure}{0.95\columnwidth}
        \includegraphics[width=\linewidth]{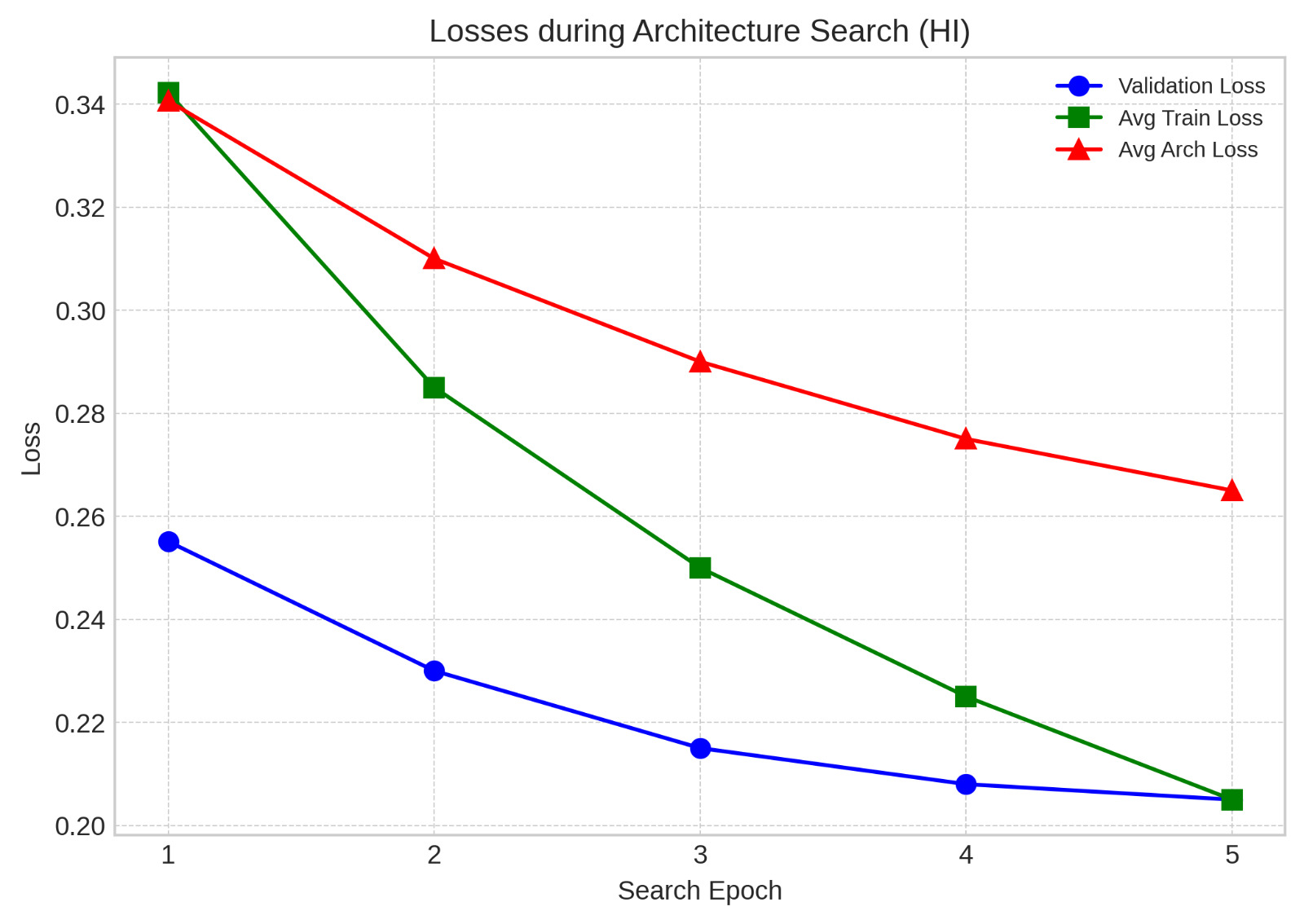}
        \caption{Training, architecture, and validation losses during the search for Hindi, showing smooth convergence.}
        \label{fig:losses_hi}
    \end{subfigure}
    \caption{Architecture search results for Hindi.}
    \label{fig:hindi_results}
\end{figure}

The validation performance plots for Hindi (Figure~\ref{fig:hindi_results}) visually confirm the effectiveness of the search. The F1-score and accuracy curves show a clear upward trend, while all three loss trajectories—validation, train, and architecture—descend smoothly, indicating an efficient search process.

\begin{figure*}[!t]
    \centering
    \includegraphics[width=\textwidth]{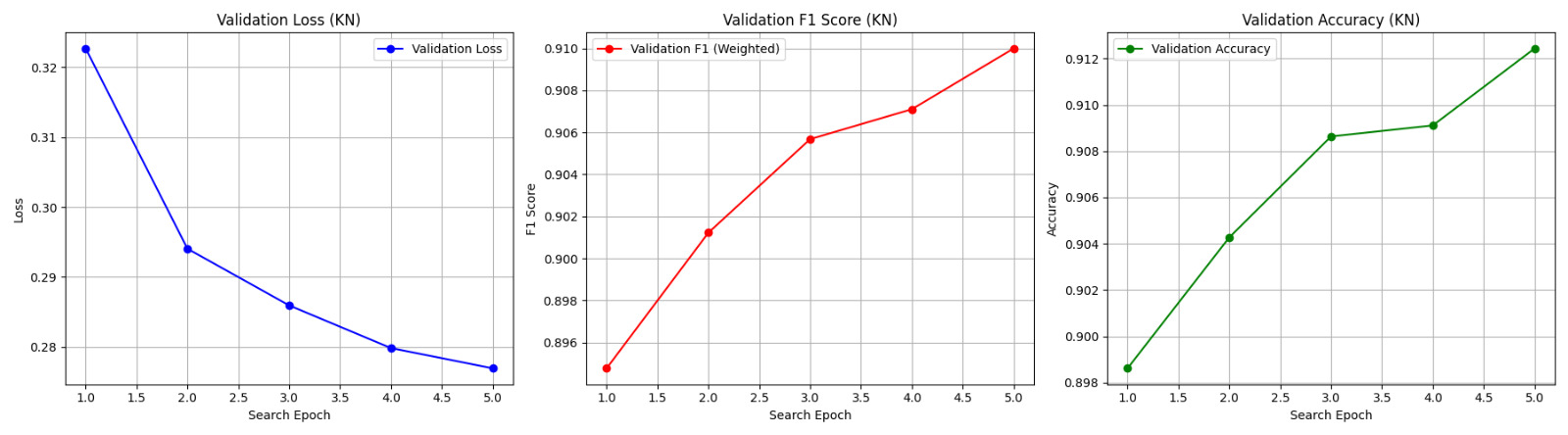}
    \caption{Validation metrics during the architecture search for Kannada (KN). From left to right: Validation Loss, Validation F1 Score, and Validation Accuracy.}
    \label{fig:kannada_plots}
\end{figure*}

The final test evaluation for Kannada (Table~\ref{tab:kannada_results}) demonstrates robust performance on unseen data. The model achieved a low test loss of 0.2031, a high test accuracy of 0.9354, and a strong weighted F1-score of 0.9331. The validation metrics during the search for Kannada (Figure~\ref{fig:kannada_plots}) also reflect successful optimization, with the loss decreasing while accuracy and F1-score consistently increased.

\begin{table*}[!t]
    \centering
    \caption{Final Test Results for Kannada After Training the Discovered Architecture.}
    \label{tab:kannada_results}
    \begin{subtable}{0.35\linewidth}
        \centering
        \caption{Overall Test Metrics}
        \begin{tabular}{@{}lr@{}}
            \toprule
            \textbf{Metric} & \textbf{Value} \\
            \midrule
            Test Loss & 0.2031 \\
            Test Accuracy & 0.9354 \\
            Test F1 (Macro) & 0.8115 \\
            Test F1 (Micro) & 0.9354 \\
            Test F1 (Weighted) & 0.9331 \\
            \bottomrule
        \end{tabular}
    \end{subtable}%
    \hfill
    \begin{subtable}{0.6\linewidth}
        \centering
        \caption{Per-Class Classification Report}
        \begin{tabular}{@{}lrrrr@{}}
            \toprule
            \textbf{Class} & \textbf{Precision} & \textbf{Recall} & \textbf{F1-Score} & \textbf{Support} \\
            \midrule
            B-LOC & 0.91 & 0.81 & 0.86 & 397 \\
            B-ORG & 0.74 & 0.78 & 0.76 & 291 \\
            B-PER & 0.90 & 0.89 & 0.89 & 614 \\
            I-LOC & 0.91 & 0.42 & 0.57 & 147 \\
            I-ORG & 0.75 & 0.67 & 0.71 & 251 \\
            I-PER & 0.95 & 0.91 & 0.93 & 527 \\
            O & 0.95 & 0.98 & 0.96 & 6901 \\
            \midrule
            Macro Avg & 0.87 & 0.78 & 0.81 & 9128 \\
            Weighted Avg & 0.93 & 0.94 & 0.93 & 9128 \\
            \bottomrule
        \end{tabular}
    \end{subtable}
\end{table*}

\section{Conclusion}
In this paper, we presented MorphNAS, a differentiable neural architecture search framework tailored to the challenges of NER in morphologically rich languages. By enhancing the DARTS approach with linguistic awareness, MorphNAS successfully automates the design of micro-architectural blocks within a searchable cell-based architecture. Our initial results on Hindi and Kannada demonstrate the framework's ability to discover effective, customized architectures. MorphNAS represents a significant step towards automating the creation of linguistically-sensitive NLP models, offering an efficient and scalable alternative to manual design.

\section{Future Work}
While MorphNAS has shown great promise, several avenues for future work exist. We plan to extend the framework to a wider range of languages and other NLP tasks, such as Machine Translation and Part-of-Speech tagging. The search space could be enriched with more sophisticated, linguistically-motivated operations. Exploring recent advancements in NAS, such as second-order DARTS variants, could further improve performance. Additionally, investigating transfer learning of discovered architectures across related languages is a promising direction for future research.

\clearpage

\bibliographystyle{plain}
\bibliography{main}

\begin{thebibliography}{1}

\bibitem{Chen2020}
Yi-Chen Chen, Jui-Yang Hsu, Cheng-Kuang Lee, and Hung-yi Lee.
\newblock {DARTS-ASR: Differentiable architecture search for multilingual speech recognition and adaptation}.
\newblock {\em arXiv preprint arXiv:2005.07029}, 2020.

\bibitem{Hundt2019}
Andrew Hundt, Varun Jain, and Gregory~D. Hager.
\newblock {sharpdarts: Faster and more accurate differentiable architecture search}.
\newblock {\em arXiv preprint arXiv:1903.09900}, 2019.

\bibitem{Jiang2019}
Yufan Jiang, Chi Hu, Tong Xiao, Chunliang Zhang, and Jingbo Zhu.
\newblock {Improved differentiable architecture search for language modeling and named entity recognition}.
\newblock In {\em Proceedings of the 2019 Conference on Empirical Methods in Natural Language Processing and the 9th International Joint Conference on Natural Language Processing (EMNLP-IJCNLP)}, pages 3585--3590, 2019.

\bibitem{Jie2021}
Renlong Jie and Junbin Gao.
\newblock {Differentiable Neural Architecture Search with Morphism-based Transformable Backbone Architectures}.
\newblock {\em arXiv preprint arXiv:2106.07211}, 2021.

\bibitem{Liu2018}
Hanxiao Liu, Karen Simonyan, and Yiming Yang.
\newblock {Darts: Differentiable architecture search}.
\newblock {\em arXiv preprint arXiv:1806.09055}, 2018.

\end{thebibliography}

\end{document}